\documentclass{article}

\usepackage{microtype}
\usepackage{graphicx}
\usepackage{subcaption}
\usepackage{booktabs}
\usepackage{hyperref}

\usepackage[preprint]{icml2026}
\usepackage{amsmath}
\usepackage{amssymb}
\usepackage{mathtools}
\usepackage{amsthm}
\usepackage[capitalize,noabbrev]{cleveref}
\usepackage[textsize=tiny]{todonotes}
\usepackage{float}

\icmltitlerunning{Physics-Informed Policy Optimization via Analytic Dynamics Regularization}

\begin{document}

\twocolumn[
  \icmltitle{Physics-Informed Policy Optimization via Analytic Dynamics Regularization}

  \icmlsetsymbol{equal}{*}

  \begin{icmlauthorlist}
    \icmlauthor{Namai Chandra}{yyy}
    \icmlauthor{Liu Mohan}{comp}
    \icmlauthor{Zhihao Gu}{comp}
    \icmlauthor{Lin Wang$^{\dag}$}{comp}
  
  \end{icmlauthorlist}

  \icmlaffiliation{yyy}{Electronic Systems, IIT Madras, India}
  \icmlaffiliation{comp}{EmPACT Lab, Nanyang Technological University, Singapore}
  \icmlcorrespondingauthor{Namai Chandra (work done during intern at NTU EmPACT Lab)}{23f3000200@es.study.iitm.ac.in}
  \icmlcorrespondingauthor{Lin Wang} {linwang@ntu.edu.sg}

  \vskip 0.3in
]
\printAffiliationsAndNotice{}
% \begin{abstract}
% Robotic control policies learned purely from data often suffer from high sample complexity and physically inconsistent behavior. In contrast, robotic systems are governed by well-defined mechanical laws that are readily available through their simulation models. We propose a physics-informed reinforcement learning framework that explicitly integrates robot dynamics into policy optimization by combining Physics-Informed Neural Networks (PINNs) with policy-gradient methods.

% Starting from a robot’s MuJoCo XML description, we automatically extract kinematic structure, joint parameters, and inertial properties to construct the Lagrangian dynamics model of the manipulator. These physical quantities are used to enforce the equations of motion as a differentiable constraint during learning. A neural network predicts joint accelerations, while deviations from the Lagrangian dynamics are penalized through a physics residual term that regularizes the policy update.

% By embedding known physics directly into the learning process, the proposed approach biases neural policies toward dynamically consistent solutions, improving learning efficiency, stability, and control accuracy. This formulation is general, model-agnostic, and applicable to single- and multi-manipulator systems without requiring task-specific redesign of the learning algorithm.
% \end{abstract}

\begin{abstract}
Reinforcement learning (RL) has achieved strong performance in robotic control; however, state-of-the-art policy learning methods, such as actor-critic methods, still suffer from high sample complexity and often produce physically inconsistent actions. This limitation stems from neural policies implicitly rediscovering complex physics from data alone, despite accurate dynamics models being readily available in simulators.
In this paper, we introduce a novel physics-informed RL framework, called \textbf{PIPER}, that seamlessly integrates physical constraints directly into neural policy optimization with analytical \textit{soft physics constraints}. At the core of our method is the integration of a \textbf{differentiable Lagrangian residual} as a regularization term within the actor's objective. This residual, extracted from a robot’s simulator description, subtly biases policy updates towards dynamically consistent solutions. Crucially, \textit{this physics integration is realized through an additional loss term during policy optimization, requiring no alterations to existing simulators or core RL algorithms}.
Extensive experiments demonstrate that our method significantly improves learning efficiency, stability, and control accuracy, establishing a \textit{new} paradigm for efficient and physically consistent robotic control.

\end{abstract}

\vspace{-20pt}
\section{Introduction}
Reinforcement learning (RL) has achieved strong performance in robotic control, establishing itself as a dominant paradigm for solving complex manipulation and locomotion tasks. State-of-the-art actor-critic algorithms, most notably Proximal Policy Optimization (PPO) \cite{schulman2017ppo}, Soft Actor-Critic (SAC) \cite{haarnoja2018sac}, and Twin Delayed Deep Deterministic Policy Gradient (TD3) \cite{fujimoto2018td3}, have demonstrated the ability to learn sophisticated behaviors directly from high-dimensional sensor observations. However, despite these empirical successes, data-driven RL still contends with high sample complexity and often produces physically inconsistent actions. This persistent limitation raises a critical research question: \textit{How can we bridge the gap between data-driven policy learning and the well-defined physical laws that govern robotic systems?}
\begin{figure}[t!]
    \centering
    \includegraphics[width=1.0\linewidth]{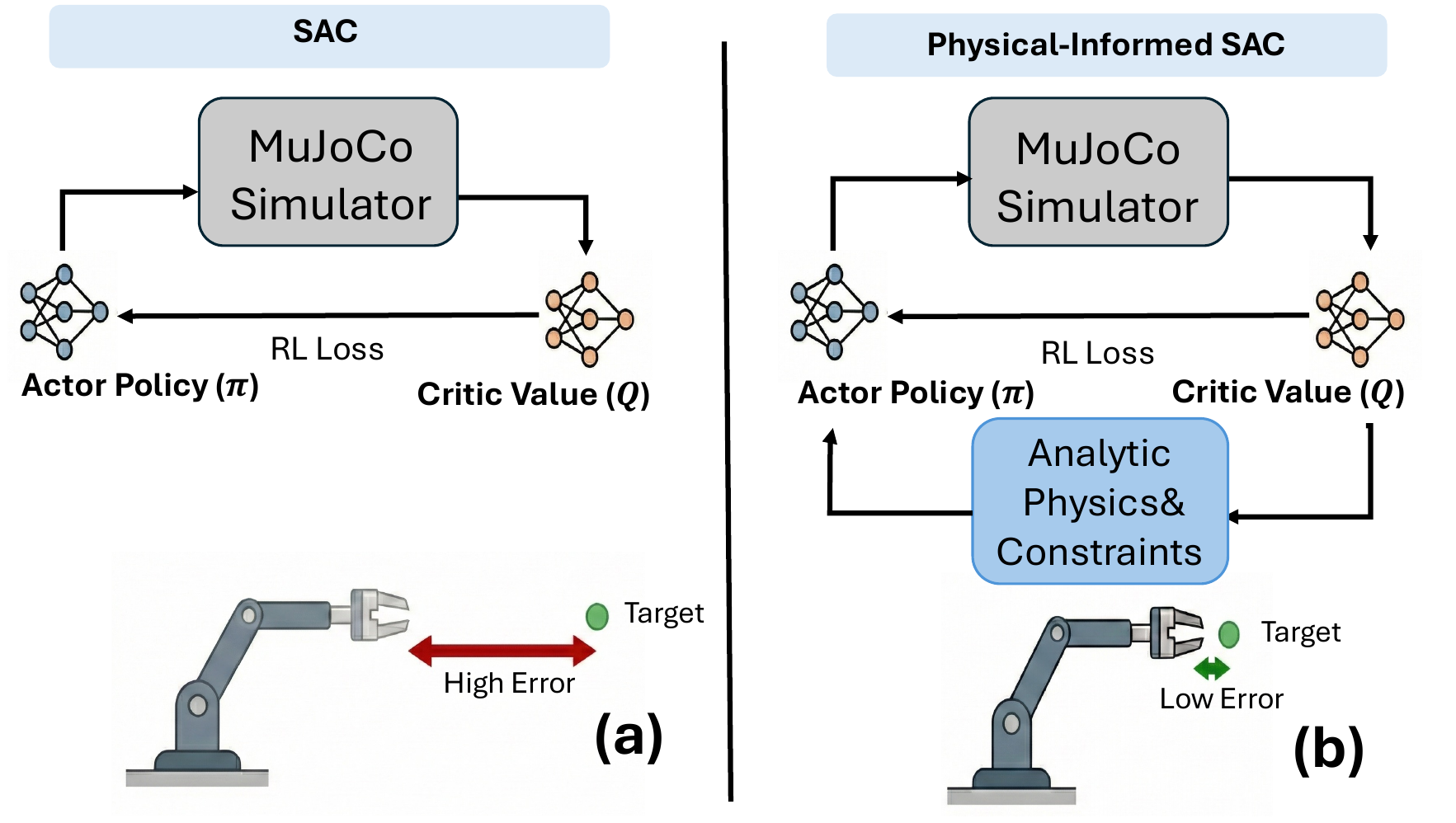}
      \vspace{-15pt}
    \caption{\textbf{Comparison of architecture and performance between SAC and Physical-Informed SAC}. (a) The standard SAC framework relying solely on the MuJoCo simulator, resulting in High Error in the robotic arm task. (b) The Physical-Informed SAC framework which incorporates an Analytic Physics \& Constraints module, leading to Low Error and precise control.}
    \vspace{-10pt}
    \label{fig:head}
\end{figure}

This limitation stems from neural policies implicitly rediscovering complex physics from data alone, despite accurate dynamics models being readily available in simulators. Standard model-free RL treats the environment as a "black box", mapping states to rewards without leveraging the underlying mechanical structure. This \textit{tabula rasa} approach forces the agent to spend millions of timesteps relearning basic concepts such as inertia, gravity, and momentum conservation via trial-and-error. As illustrated in Fig.~\ref{fig:head}, policies learned in this manner often exhibit high-frequency oscillation or "jitter", as the neural network exploits simulator integration errors rather than learning smooth, energy-efficient motion. While a modern physics engine like MuJoCo \cite{todorov2012mujoco} explicitly encodes Euler-Lagrange dynamics ($M, C, G$), standard algorithms discard this rich information during optimization.

\begin{figure}[t!]
    \centering
    \includegraphics[width=1\columnwidth]{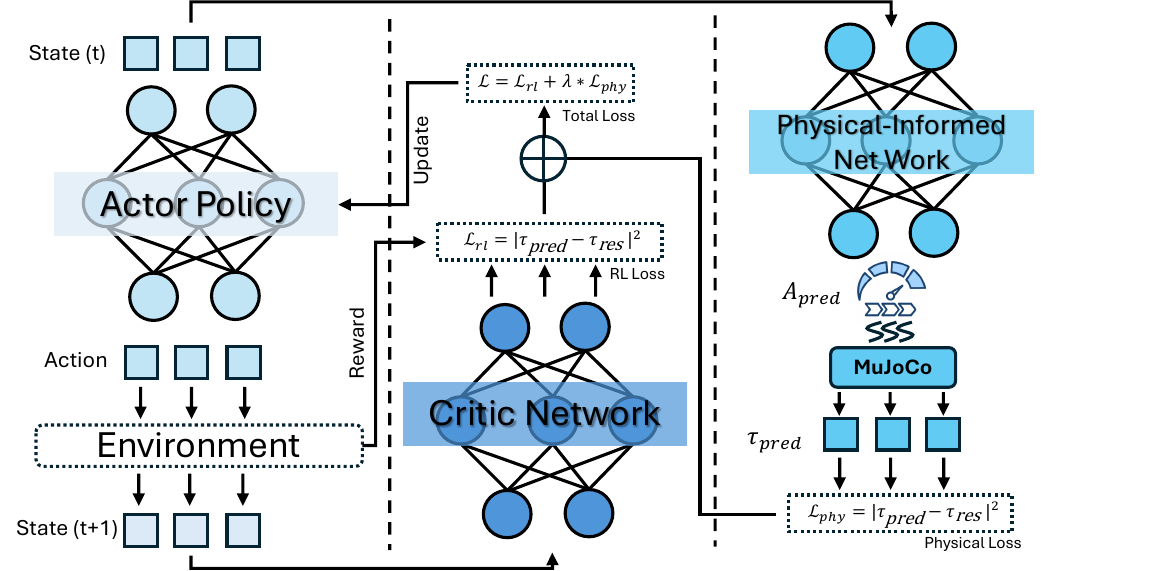}
    \caption{
    \textbf{Overview of the physics-informed actor–critic reinforcement learning framework.}
    The actor policy generates control actions based on the current system state, while the critic network evaluates action quality by estimating the value function. A PINN, coupled with a high-fidelity simulator (MuJoCo), embeds prior physical laws and system dynamics into the learning loop, providing physically consistent state transitions and constraints. The interaction between data-driven policy optimization and physics-based modeling improves training stability, sample efficiency, and generalization performance compared to purely data-driven reinforcement learning approaches.
    }
    \vspace{-15pt}
    \label{fig:pipeline}
\end{figure}

In this paper, we introduce \textbf{PIPER} (Physics-Informed Policy Optimization with Analytical Dynamic Regularization), a novel framework that seamlessly integrates physical constraints directly into neural policy optimization. By "analytical," we refer to the explicit utilization of the simulator's internal physics dynamics file to derive governing constraints, employing them as \textit{soft physics constraints} to ground the policy in the system's true mechanics. Our key idea is \textit{to move beyond the black-box assumption by providing the policy with a dense, structure-aware gradient that complements the sparse reward}. This formulation is inspired by recent advances in physics-informed learning \cite{raissi2019pinn, liu2024pinnrobots} and structured architectures \cite{greydanus2019hamiltonian}, but addresses the open challenge of integrating such priors into model-free policy gradients without requiring rigid safety layers \cite{dalal2018safe} or handcrafted controllers \cite{johannink2019residual}.

At the core of our method is the integration of a \textbf{differentiable Lagrangian residual} as a regularization term within the actor's objective. By leveraging the exact Euler-Lagrange terms ($M, C, G$) extracted from the robot’s simulator description at runtime, we construct a residual vector that quantifies the precise violation of the equations of motion for any given action. This residual subtly biases policy updates towards dynamically consistent solutions. Crucially, \textit{this physics integration is realized through an additional loss term during policy optimization, requiring no alterations to existing simulators or core RL algorithms}. This mechanism acts as a differentiable "physics coach," guiding the policy search manifold toward feasible regions without enforcing hard constraints that could hinder exploration.

Extensive experiments demonstrate that our method significantly improves learning efficiency, stability, and control accuracy. We validate our approach on a curriculum of robotic tasks ranging from kinematic reaching to complex object manipulation. Our results show substantial gains across all metrics: success rates improve by over 7\% in contact-rich tasks (FetchSlide), final precision improves by up to 79.5\% (PIPER-TD3), learning stability increases by 42\% (PickAndPlace), and sample efficiency improves by over 45\% (PIPER-SAC). By grounding the policy search in analytic mechanics, PIPER establishes a new paradigm for efficient and physically consistent robotic control.
In summary, our contributions are as follows:
\begin{itemize}
    \item \textbf{Physics-Informed Policy Optimization with Explicit Residuals (PIPER):} We introduce a framework that regularizes reinforcement learning using analytic dynamics. By minimizing a differentiable Lagrangian residual, we enforce physical consistency directly within the policy optimization landscape.
    \item \textbf{Plug-and-Play Dynamics Extraction:} We propose an ``Dynamics Oracle" that extracts exact inertial properties ($M, C, G$) from  MuJoCo XML descriptors at runtime, enabling physics-aware learning on arbitrary kinematic chains without manual derivation.
    \item \textbf{Algorithm-Agnostic Applicability:} We demonstrate that PIPER is compatible with a wide range of algorithms, including on-policy (PPO), off-policy (TD3), and distributional methods (SAC, TQC), without modifying their core update rules.
    \item \textbf{Robustness in Contact-Rich Regimes:} We validate our approach on a curriculum of robotic tasks ranging from reaching to complex object manipulation (Push, Slide). Our results show that physics regularization significantly improves sample efficiency and asymptotic precision, outperforming state-of-the-art baselines in sparse-reward settings.
\end{itemize}

\vspace{-10pt}
\section{Related Works}

\textbf{Model-Free Actor--Critic Reinforcement Learning.}
Model-free RL has become the dominant paradigm for continuous control in robotics. Early actor–critic algorithms such as Deep Deterministic Policy Gradient (DDPG)~\cite{lillicrap2019continuouscontroldeepreinforcement} enabled learning of continuous control policies directly from high-dimensional observations, but suffered from instability and overestimation bias. Proximal Policy Optimization (PPO)~\cite{schulman2017ppo} stabilized on-policy updates through a clipped surrogate objective, making it one of the most widely adopted algorithms in robotics and simulation. Twin Delayed Deep Deterministic Policy Gradient (TD3)~\cite{fujimoto2018td3} reduced overestimation errors in off-policy learning by employing twin Q-networks and delayed policy updates. Soft Actor–Critic (SAC)~\cite{haarnoja2018sac} further improved sample efficiency and exploration by incorporating entropy regularization into the actor–critic framework. Despite their empirical success, these model-free algorithms treat the environment as a black box and ignore known physical structures of robotic systems. Recent efforts have sought to improve optimization stability through conservative updates~\cite{garg2023extreme, kostrikov2022iql} or normalization choices~\cite{islam2023revisiting}. \textit{In contrast to these data-driven approaches, PIPER explicitly injects known physical structures into the policy gradient update via Lagrangian regularization, preventing the policy from violating fundamental laws of motion during exploration.}

\textbf{Incorporating Physical Priors into Learning-Based Control}
A growing line of work incorporates physical priors into learning-based control to improve stability, safety, and data efficiency. Early approaches focused on enforcing constraints during policy optimization, such as Constrained Policy Optimization (CPO), which integrates trust-region updates with explicit safety constraints~\cite{achiam2017cpo}. Complementary methods introduce lightweight safety mechanisms, including action correction layers~\cite{dalal2018safe} and residual reinforcement learning, which augments classical controllers with learned residuals~\cite{silver2018residual, johannink2019residual}. More recently, inductive biases have explored energy conservation~\cite{zanella2024energytanks} and symmetry-aware representations~\cite{zhao2024equivariant}. \textit{Unlike these approaches, our framework avoids the inference-time latency of solving quadratic programs for hard constraints and does not require a handcrafted base controller. Furthermore, while methods like Deep Lagrangian Networks \cite{lutter2019deeplagrangian} explicitly embed physics into model-based learning, PIPER enforces physics as a soft, differentiable manifold constraint directly within the model-free policy learning objective.}

\textbf{Physics-Informed Neural Networks (PINNs) for Dynamics and Control.}
PINNs have emerged as a powerful framework for integrating physical laws into neural models by enforcing governing equations as soft constraints~\cite{raissi2019pinn}. In robotics, PINNs are primarily used to learn forward or inverse dynamics models for system identification~\cite{liu2024pinnrobots, liu2022physicsaware}. Several works leverage PINNs as surrogate dynamics models within reinforcement learning, where physics residuals regularize trajectory prediction or state transitions~\cite{chen2022pinnrl, tang2023physicsrl}. However, these methods typically rely on \textit{approximating} the physical laws from data, which introduces epistemic uncertainty. Related approaches incorporate analytic dynamics into model-based RL, but they often require computationally expensive rollouts or learned world models to propagate physical consistency~\cite{deisenroth2013pilco, janner2019mbpo}. \textit{We diverge from this paradigm by utilizing an exact Analytic Dynamics Oracle that extracts ground-truth inertial properties directly from the simulator. This allows us to use the PINN solely as a training-time `physics coach' that regularizes the actor directly, ensuring the final policy remains lightweight and deployment-ready without requiring a forward dynamics model or online planning.}

\begin{table*}[t]
\centering
\caption{Comparison of physics-aware reinforcement learning approaches.}
\vspace{-5pt}
\label{tab:positioning}
\resizebox{0.95\textwidth}{!}{%
\begin{tabular}{lcccc}
\toprule
Method & 
\textbf{Model-Free} & 
\textbf{No Test-Time Optimization} & 
\textbf{Algorithm Agnostic} & 
\textbf{Uses Lagrangian Structure} \\
\midrule
PPO / SAC / TD3 & 
\textcolor{green!30!black}{\checkmark} & 
\textcolor{green!30!black}{\checkmark} & 
\textcolor{green!30!black}{\checkmark} & 
\textcolor{red}{\texttimes} \\

Deep Lagrangian \cite{lutter2019deeplagrangian} & 
\textcolor{red}{\texttimes} & 
\textcolor{green!30!black}{\checkmark} & 
\textcolor{red}{\texttimes} & 
\textcolor{green!30!black}{\checkmark} \\

Residual RL \cite{johannink2019residual} & 
\textcolor{green!30!black}{\checkmark} & 
\textcolor{green!30!black}{\checkmark} & 
\textcolor{green!30!black}{\checkmark} & 
\textcolor{red}{\texttimes} \\

Safety Layer / CPO \cite{dalal2018safe} & 
\textcolor{green!30!black}{\checkmark} & 
\textcolor{red}{\texttimes} & 
\textcolor{red}{\texttimes} & 
\textcolor{red}{\texttimes} \\

HJB-RL \cite{cheng2021hjbrl} & 
\textcolor{green!30!black}{\checkmark} & 
\textcolor{red}{\texttimes} & 
\textcolor{red}{\texttimes} & 
\textcolor{red}{\texttimes} \\

PI-MBRL \cite{pmlr-v211-ramesh23a} & 
\textcolor{red}{\texttimes} & 
\textcolor{green!30!black}{\checkmark} & 
\textcolor{red}{\texttimes} & 
\textcolor{red}{\texttimes} \\

\textbf{Ours (PIPER)} & 
\textbf{\textcolor{green!30!black}{\checkmark}} & 
\textbf{\textcolor{green!30!black}{\checkmark}} & 
\textbf{\textcolor{green!30!black}{\checkmark}} & 
\textbf{\textcolor{green!30!black}{\checkmark}} \\
\bottomrule
\end{tabular}
}
\end{table*}

\vspace{-10pt}
\section{Methodology}
\label{sec:method}

We propose \textbf{PIPER} (Physics-Informed Policy Optimization), a framework that regularizes policy learning by explicitly enforcing the laws of rigid-body mechanics. In this section, we first formalize the underlying physical principles and the reinforcement learning setting, then detail the construction of the differentiable Lagrangian residual and its integration into the policy gradient update.

\vspace{-5pt}
\subsection{Preliminaries}
\vspace{-5pt}

\textbf{Lagrangian Rigid-Body Mechanics.}
We model the manipulator as a holonomic kinematic chain with generalized coordinates $q \in \mathbb{R}^n$ and velocities $\dot{q} \in \mathbb{R}^n$. Its dynamics are governed by the Principle of Least Action, derived from the Lagrangian $\mathcal{L}: T\mathcal{Q} \to \mathbb{R}$, defined as the difference between kinetic energy $\mathcal{K}$ and potential energy $\mathcal{P}$:
\begin{align}
    \mathcal{L}(q, \dot{q}) &= \mathcal{K}(q, \dot{q}) - \mathcal{P}(q) \nonumber \\
    &= \frac{1}{2} \dot{q}^T M(q) \dot{q} - \mathcal{P}(q).
\end{align}
The equations of motion are obtained by applying the Euler-Lagrange operator $\frac{d}{dt}(\frac{\partial \mathcal{L}}{\partial \dot{q}}) - \frac{\partial \mathcal{L}}{\partial q} = \tau$. This yields the explicit structural form:
\begin{equation}
    \underbrace{M(q)\ddot{q}}_{\text{Inertial}} + \underbrace{C(q,\dot{q})\dot{q}}_{\text{Coriolis/Centrifugal}} + \underbrace{G(q)}_{\text{Gravitational}} = \tau + \tau_{\text{ext}}.
    \label{eq:el_full}
\end{equation}
Here, $M(q) \in \mathbb{S}_{++}^n$ is the symmetric positive-definite inertia matrix, and $C(q,\dot{q})$ is the Coriolis matrix defined by Christoffel symbols of the first kind $c_{ijk} = \frac{1}{2}(\frac{\partial M_{kj}}{\partial q_i} + \frac{\partial M_{ki}}{\partial q_j} - \frac{\partial M_{ij}}{\partial q_k})$:
\begin{equation}
    C_{kj}(q, \dot{q}) = \sum_{i=1}^{n} c_{ijk}(q) \dot{q}_i.  
\end{equation}
For efficient computation in our framework, we define the \textit{Generalized Bias Force} $\mathbf{b}(s)$, which aggregates all state-dependent forces excluding inertial acceleration:
\begin{equation}
    \mathbf{b}(q, \dot{q}) \triangleq C(q,\dot{q})\dot{q} + G(q) - \tau_{\text{ext}}.
    \label{eq:bias_def}
\end{equation}
Substituting Eq.~\eqref{eq:bias_def} into Eq.~\eqref{eq:el_full} allows us to express the system dynamics in the compact \textit{Inverse Dynamics} form:
\begin{equation}
    \mathcal{ID}(s, \ddot{q}) \mapsto \tau \implies M(q)\ddot{q} + \mathbf{b}(s) = \tau.
    \label{eq:inverse_dynamics_compact}
\end{equation}

\textbf{Action Space and Goal-Conditioned Distributional RL.}\\ We assume an action space $\mathcal{A}$ corresponding to direct joint torque control, such that the policy outputs proposed torques $a = \tau_\theta$.
We operate within a Goal-Conditioned Markov Decision Process (GC-MDP) tuple $(\mathcal{S}, \mathcal{A}, P, R, \gamma, \mathcal{G})$. To mitigate overestimation bias, we employ Truncated Quantile Critics (TQC) \cite{kuznetsov2020tqc}, where the return distribution $Z$ is approximated by a set of quantiles minimizing the Huber loss. Our framework integrates physics constraints directly into this loop, ensuring the learned policy remains consistent with Eq.~\eqref{eq:inverse_dynamics_compact}.

\subsection{Automated Dynamics Oracle (ADO)}
To enforce Eq.~\eqref{eq:inverse_dynamics_compact} without manual derivation, we implement a \textit{Dynamics Oracle} that queries the simulator's internal spatial algebra engines at runtime.
The inertia matrix $M(q)$ is computed via the Composite Rigid Body Algorithm (CRBA)~\cite{featherstone2014rigid}, which projects spatial inertias $\mathcal{I}_i$ onto the joint motion subspace $S_i$:
\begin{equation}
    M_{ij}(q) = \text{CRBA}(q) = \sum_{k \in \text{subtree}(i)} S_i^T {}^kX_i^T \mathcal{I}_k {}^kX_j S_j.
\end{equation}
Simultaneously, the generalized bias $\mathbf{b}(q, \dot{q})$ is extracted via the Recursive Newton-Euler Algorithm (RNEA) by creating a "zero-acceleration" virtual state ($\ddot{q}=\mathbf{0}$):
\begin{equation}
    \mathbf{b}(q, \dot{q}) = \text{RNEA}(q, \dot{q}, \ddot{q}=\mathbf{0}, \mathbf{g}, \mathbf{f}_{\text{ext}}) = C(q,\dot{q})\dot{q} + G(q) - \tau_{\text{ext}}. \end{equation} External contact forces $\tau_{\text{ext}}$ are directly queried from MuJoCo's contact sensors at runtime to ensure the bias vector remains accurate in contact-rich regimes. These quantities provide ground-truth physical terms for regularization, decoupled from the neural network's learning process.

\subsection{Physics-Informed Policy Optimization}
Since the instantaneous acceleration $\ddot{q}$ is not directly observable, we introduce a proxy network $\Phi_\phi$ (PINN) to predict joint accelerations $\hat{\ddot{q}}_t = \Phi_\phi(s_t, a_t)$. The PINN is trained to minimize the prediction error against finite-difference accelerations observed in the environment:
\begin{equation}
    \mathcal{L}_{\text{PINN}}(\phi) = \mathbb{E}_{\mathcal{D}} \left[ \| \Phi_\phi(s, a) - \ddot{q}^{\text{obs}} \|_2^2 + \beta \| \mathbf{r}(s, a) \|_2^2 \right].
\end{equation}

\textbf{Lagrangian Residual.}
We quantify physical inconsistency by defining the \textit{Physics Residual} vector $\mathbf{r}(s, a)$. By substituting the PINN prediction and the Oracle terms into the inverse dynamics formulation (Eq.~\eqref{eq:inverse_dynamics_compact}), we obtain:
\begin{align}
    \hat{\tau}_{\text{phy}}(s, a) &= M(q)\Phi_\phi(s, a) + \mathbf{b}(s), \\
    \mathbf{r}(s, a) &= \hat{\tau}_{\text{phy}}(s, a) - a,
    \label{eq:residual_def}
\end{align}
where $a$ corresponds to the joint torques proposed by the policy. This formulation ensures that the physics penalty evaluates the physical consistency of the current policy's chosen action, preventing it from collapsing into the PINN's mere prediction error and resolving off-policy validity issues.

\textbf{Policy Regularization.}
This residual is incorporated into the policy optimization via a weighted penalty term $\mathcal{L}_{\text{phys}}$. The augmented actor objective becomes:
\begin{equation}
    \max_\theta \mathcal{J}_{\text{total}} = \underbrace{\mathbb{E}[\min(\rho_t A_t, \text{clip}(\dots)A_t)]}_{\mathcal{L}_{\text{PPO}}} - \lambda_{\text{phys}} \underbrace{\mathbb{E}[ \| \mathbf{r}(s, \pi_\theta(s)) \|_2^2 ]}_{\mathcal{L}_{\text{phys}}}.
    \label{eq:augmented_objective}
\end{equation}

\textbf{Gradient Flow and Optimization.}
The physics loss provides an auxiliary gradient signal that shapes the policy manifold. By applying the chain rule to $\mathcal{L}_{\text{phys}}$ (Eq.~\eqref{eq:augmented_objective}), we derive the explicit update direction:
\begin{equation}
    \nabla_\theta \mathcal{L}_{\text{phys}} = \mathbb{E} \left[ 2\mathbf{r}^T \left( M(q) \nabla_a \Phi_\phi \right) \nabla_\theta \pi_\theta(s) \right].
    \label{eq:grad_flow}
\end{equation}
This term effectively creates a "Physics Preconditioner" $\mathcal{H} = M \nabla_a \Phi$. It rotates the standard policy gradient to ensure that updates to the action $a$ align with the robot's inertial properties, penalizing actions that would require physically impossible changes in momentum.

\textbf{Energy Conservation Enforcement.}
Beyond instantaneous acceleration consistency, physical realism requires adherence to the First Law of Thermodynamics. For a conservative system, the time derivative of the total mechanical energy $E(q, \dot{q}) = \mathcal{K}(q, \dot{q}) + \mathcal{P}(q)$ must equal the power injected by control torques:
\begin{equation}
    \frac{dE}{dt} = \dot{q}^T \tau.
\end{equation}
Using the skew-symmetric property of $\dot{M} - 2C$ (where $\dot{M}$ is approximated via finite differences), we derive the explicit energy violation residual:
\begin{equation}
    r_{\text{energy}} = \left| \dot{q}^T M(q) \hat{\ddot{q}} + \frac{1}{2}\dot{q}^T \dot{M}(q) \dot{q} + \dot{q}^T G(q) - \dot{q}^T \tau \right|.
\end{equation}
We incorporate this term into the PINN loss for contact-rich tasks (Push/Slide) where energy dissipation via friction must be explicitly modeled. This ensures that the learned dynamics do not generate "free energy," preventing the policy from exploiting simulator integration errors.

\subsection{Task-Specific Constraint Instantiation}
While the general Lagrangian residual (Eq.~\ref{eq:residual_def}) captures fundamentally rigid-body dynamics, specific manipulation regimes require tailored constraint formulations to account for contact forces and hybrid dynamics. We instantiate the physics loss $\mathcal{L}_{\text{phys}}$ for four distinct control regimes:

\textbf{Kinematic Control (Free-Space).} For reaching tasks where external contact is negligible, we focus strictly on internal articulation consistency and kinematic feasibility:
\begin{equation}
    \mathcal{L}_{\text{reach}} = \lambda_1 \| M(q)\hat{\ddot{q}} + \mathbf{b}(q,\dot{q}) - \tau \|_2^2 + \lambda_2 \| \phi_{\text{FK}}(q) - g \|_2^2,
\end{equation}
where $\phi_{\text{FK}}$ denotes the forward kinematics function and $g$ is the target coordinates.

\textbf{Contact Dynamics (Continuous Force).} For pushing tasks involving friction, we augment the loss to enforce the work-energy theorem explicitly. The residual includes a friction dissipation term:
\begin{equation}
    \mathcal{L}_{\text{push}} = \mathcal{L}_{\text{reach}} + \lambda_f \left| \underbrace{\int \mu |F_n| \|v_{obj}\| dt}_{\text{Work done by friction}} - \Delta E_{\text{kinetic}} \right|.
    \label{eq:push_constraint}
\end{equation}
This penalizes violations of energy conservation, preventing the policy from exploiting simulator inaccuracies to generate unrealistic impulses.

\textbf{Impulse Transfer (Discontinuous Force).} For sliding tasks dominated by momentum transfer ($J = \int F_{ext} dt$), we enforce the impulse-momentum theorem at the point of impact:
\begin{equation}
    \mathcal{L}_{\text{slide}} = \mathcal{L}_{\text{reach}} + \lambda_m \| m_{obj}\Delta v_{obj} - J_{\text{impact}} \|_2^2, \end{equation} where $J_{\text{impact}}$ is the impact impulse extracted from the simulator's collision properties. Additionally, we enforce a sliding friction residual $r_{\text{fric}} = m a_{obj} + \mu m g \hat{v}_{obj}$ (where $\hat{v}_{obj}$ is the unit velocity vector) to ensure the predicted trajectory adheres to Coulomb friction laws. Normal forces and grip forces for these specialized regimes are acquired directly from the simulator's physics state.

\textbf{Hybrid Dynamics (Grasping).} The pick-and-place tasks introduce a hybrid dynamical system where the effective inertia changes discretely. We model this via a \textit{Combined Mass Matrix, formulated as}:
\begin{equation}
    M_{\text{comb}}(q) = M_{\text{arm}}(q) + J^T(q) m_{obj} J(q).
\end{equation}
We enforce grasp stability via a force-closure constraint, penalizing actions that attempt to lift the object without sufficient normal force to counteract gravity:
\begin{equation}
    \mathcal{L}_{\text{grasp}} = \lambda_g \max(0, m_{obj}(g + \ddot{z}_{lift}) - \mu_{\text{grip}} F_{\text{grip}})^2.
\end{equation}

\textbf{Algorithm.}
The training pipeline, defined formally as the \textit{PINN-RL Training Pipeline}, coordinates data collection, dynamics learning, and policy regularization. The framework is designed to be algorithm-agnostic, supporting both on-policy methods (PPO) and off-policy algorithms (SAC, TD3, TQC) by enforcing physical consistency directly within the policy optimization step.

\begin{algorithm}[tb]
   \caption{Generalized Physics-Informed Policy Optimization (PIPER)}
   \label{alg:piper}
\begin{algorithmic}
   \STATE {\bfseries Input:} Env $\mathcal{E}$, Oracle $\mathcal{O}_{dyn}$, Policy $\pi_\theta$, Critic $Q_\psi$, PINN $\Phi_\phi$, Buffer $\mathcal{D}$
   \STATE {\bfseries Initialize:} $\theta, \psi, \phi$, $\lambda_{phys}$
   \FOR{iteration $k=1$ {\bfseries to} $K$}
       \STATE \textbf{// 1. Interaction \& Dynamics Extraction}
       \STATE $a_t \sim \pi_\theta(s_t)$, $s_{t+1}, r_t \leftarrow \mathcal{E}.\text{step}(a_t)$
       \STATE Extract $M_t, \mathbf{b}_t \leftarrow \mathcal{O}_{dyn}(s_t)$ 
       \STATE Estimate $\tau_{\text{eff}} \leftarrow M_t \ddot{q}_t^{\text{obs}} + \mathbf{b}_t$
       \STATE Store transition $(s_t, a_t, r_t, s_{t+1}, M_t, \mathbf{b}_t, \tau_{\text{eff}})$ in $\mathcal{D}$
       
       \STATE \textbf{// 2. Train Dynamics Proxy (PINN)}
       \IF{update condition (PINN)}
           \STATE Sample batch $B_{phys} \sim \mathcal{D}$
           \STATE $\mathcal{L}_{\Phi} = \|\Phi_\phi(s, a) - \ddot{q}^{\text{obs}}\|^2 + \beta \|\mathbf{r}(s,a)\|^2$
           \STATE Update $\phi \leftarrow \phi - \eta \nabla_\phi \mathcal{L}_{\Phi}$
       \ENDIF
       
       \STATE \textbf{// 3. Physics-Regularized RL Update}
       \IF{update condition (Policy)}
           \STATE Sample batch $B_{RL} \sim \mathcal{D}$ (or Rollout)
           \STATE \textbf{if} On-Policy (PPO):
           \STATE \quad $\mathcal{L}_{\text{RL}} = \mathcal{L}_{\text{CLIP}}(\theta) + c_1 \mathcal{L}_{\text{VF}}(\psi)$
           \STATE \textbf{else if} Off-Policy (SAC/TQC/TD3):
           \STATE \quad $\mathcal{L}_{\text{RL}} = \mathcal{L}_{\text{Critic}}(\psi) - \alpha \mathbb{E}[\log \pi_\theta]$
           
           \STATE \textbf{Compute Physics Penalty:}
           \STATE $\mathcal{L}_{\text{phys}} = \| M(q)\Phi_\phi(s, \pi_\theta(s)) + \mathbf{b}(s) - \pi_\theta(s) \|^2$ \hfill $\triangleright$ Eq.~\ref{eq:residual_def}
           
           \STATE \textbf{Update:} $\theta \leftarrow \theta - \nabla_\theta (\mathcal{L}_{\text{RL}} + \lambda_{\text{phys}} \mathcal{L}_{\text{phys}})$ \hfill $\triangleright$ Eq.~\ref{eq:grad_flow}
       \ENDIF
   \ENDFOR
\end{algorithmic}
\end{algorithm}

\section{Experiments}
\label{sec:experiments}

\subsection{Experimental Setup}

\begin{figure*}[t]
    \centering
    \includegraphics[width=0.7\textwidth]{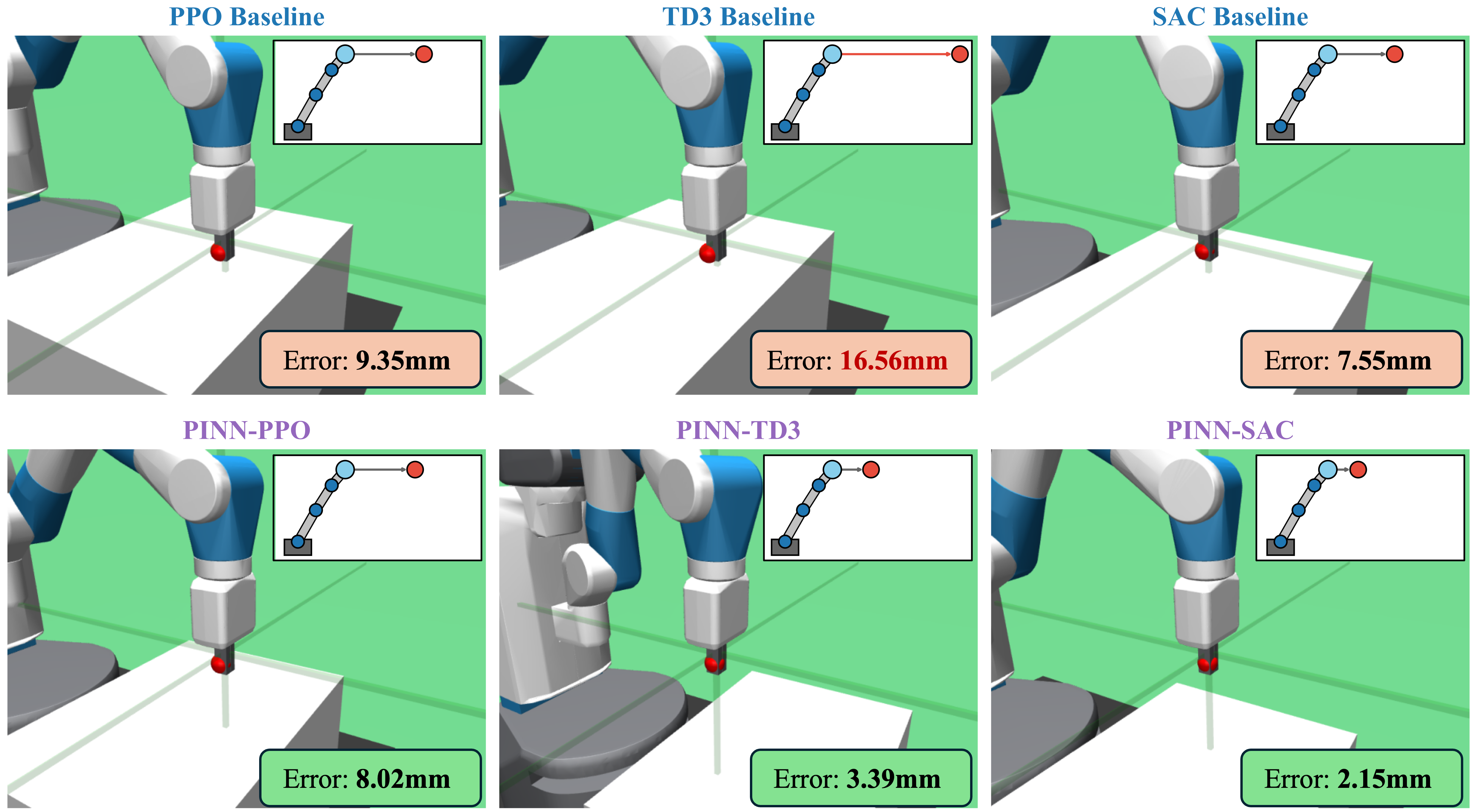}
    \caption{
    Qualitative comparison of final end-effector configurations in the FetchReach-v4 environment.
    The green background indicates the simulated MuJoCo workspace.
    Top row: Baseline policies often exhibit steady-state error or drift due to lack of structural knowledge.
    Bottom row: PIPER policies achieve tighter convergence to the target geometry by exploiting the null space of the manipulator dynamics.
    }
    \label{fig:mujoco_frames}
\end{figure*}
%why are we focusing on these 4 tasks only? they are very typical for verifying the physics
\textbf{Environments and Task Specifications.}
We evaluate PIPER on the Gymnasium-Robotics benchmark suite \cite{gymnasium}, utilizing a 7-DOF Franka Emika Panda arm. This suite consists of four distinct environments which, collectively, are exhaustive of the fundamental physics principles governing robotic manipulation, that progress in physical complexity:
\begin{itemize}
    \item \textbf{FetchReach-v4:} A pure kinematic control task with a 10-dimensional observation space ($\mathcal{S} \in \mathbb{R}^{10}$) comprising gripper position and velocity. The agent must minimize the geodesic distance to a target $g \in \mathbb{R}^3$.
    \item \textbf{FetchPush-v4:} Introduces contact mechanics. The state space expands to $\mathcal{S} \in \mathbb{R}^{25}$ to include object pose and rotational velocity. The agent must manipulate a block via planar pushing, requiring the policy to implicitly model Coulomb friction cones.
    \item \textbf{FetchSlide-v4:} A momentum-dominated task where the target is outside the robot's kinematic workspace. Success depends on imparting a precise impulse $J = \int F dt$ to slide the object across a surface with friction coefficient $\mu$.
    \item \textbf{FetchPickAndPlace-v4:} The most complex regime involving hybrid dynamics (discrete contact switching). The agent must grasp, lift, and stabilize an object, managing gravity compensation ($g=9.81 m/s^2$) and grasping constraints.
\end{itemize}

\textbf{Implementation and Hyperparameters.}
 The testbed explicitly resolves the full equations of motion for the \textbf{7-DOF Franka Emika Panda manipulator}, incorporating complex inertial coupling via the mass matrix $M(q)$ and Coriolis effects. Crucially, interaction dynamics are governed by realistic surface properties faithfully model real-world , utilizing a Coulomb friction coefficient of $\mu=0.5$ and a block mass of $0.1$ kg to challenge the policy's ability to model energy dissipation and momentum transfer. The simulation operates at a timestep of $dt=0.002s$ to capture transient contact events.

For algorithmic comparison, we utilize standard hyperparameters: PPO, TD3, and SAC use learning rates of $3 \times 10^{-4}$ to $1 \times 10^{-3}$ for \texttt{FetchReach}. For the sparse-reward manipulation tasks (\texttt{Push}, \texttt{Slide}, \texttt{PickAndPlace}), we employ TQC+HER with a replay buffer of $10^6$ transitions and a large batch size of 2048 to ensure stable gradient estimation. Our PIPER implementation augments these baselines with a Physics-Informed Neural Network (PINN) of approximately 162k parameters. The physics regularization weight $\lambda_{phys}$ is rigorously tuned to 0.01 for on-policy methods and 0.005 for off-policy methods via grid search. To ensure a fair comparison, baseline hyperparameters were subjected to an equivalent grid search optimization tailored to these specific tasks.

% \textbf{Evaluation Metrics.}
% We report results using three hardware-agnostic metrics. To ensure statistical robustness, we train \textbf{5 independent instances (random seeds)} of each algorithm. For each seed, performance is calculated as the average over \textbf{100 evaluation episodes} per checkpoint. The final results in Tables 2 and 3 report the \textbf{mean and standard deviation across these 5 seeds}.
\textbf{Evaluation Metrics.}
We report results using three hardware-agnostic metrics. To ensure both statistical robustness and reproducibility, we train \textbf{5 independent instances} of each algorithm using a fixed set of random seeds $\{42, 43, 44, 45, 46\}$. For each seed, performance is calculated as the average over \textbf{100 evaluation episodes} per checkpoint.
\begin{enumerate}
    \item \textbf{Sample Efficiency ($N_{95\%}$):} Total environment interactions required to first achieve a 95\% success rate.
    \item \textbf{Precision ($d_{final}$):} Euclidean distance to goal at termination ($d_{final} = \|p_{obj} - g\|_2$).
    \item \textbf{Stability ($\sigma$):} Standard deviation of the success rate over the final 100 evaluation rollouts (where an epoch is defined as 1,000 environment steps).
\end{enumerate}

\subsection{Results and Analysis}

\begin{table*}[tb]
\centering
\caption{Performance Metrics for \textbf{FetchReach-v4}. PIPER improves efficiency, precision, and stability across PPO, TD3, and SAC.}
\label{tab:reach_summary}
\resizebox{1.0\textwidth}{!}{%
\begin{tabular}{llccccccc}
\toprule
Algorithm & Success Rate & Final Error & Precision Gain & Stability ($\sigma$) & Stability Gain & Steps to 95\% & Efficiency Gain \\
\midrule
PPO & 100\% & 9.35 mm & -- & 10.94 & -- & 46,650 & -- \\
\textbf{PIPER-PPO} & 100\% & 8.02 mm & \textbf{\textcolor{green!60!black}{+14.2\%}} & 8.13 & \textbf{\textcolor{green!60!black}{+25.7\%}} & 32,800 & \textbf{\textcolor{green!60!black}{+30\%}} \\
\midrule
TD3 & 100\% & 16.56 mm & -- & 17.59 & -- & 23,850 & -- \\
\textbf{PIPER-TD3} & 100\% & 3.39 mm & \textbf{\textcolor{green!60!black}{+79.5\%}} & 11.53 & \textbf{\textcolor{green!60!black}{+34.5\%}} & 17,550 & \textbf{\textcolor{green!60!black}{+26\%}} \\
\midrule
SAC & 100\% & 7.55 mm & -- & 11.41 & -- & 24,250 & -- \\
\textbf{PIPER-SAC} & 100\% & \textbf{2.15 mm} & \textbf{\textcolor{green!60!black}{+71.5\%}} & \textbf{8.10} & \textbf{\textcolor{green!60!black}{+29.0\%}} & \textbf{13,250} & \textbf{\textcolor{green!60!black}{+45\%}} \\
\bottomrule
\end{tabular}
}
\end{table*}

\subsubsection{Kinematic Control (FetchReach)}
The agent operates within a holonomic kinematic chain without external contact forces. The primary challenge is generating smooth, minimum-jerk trajectories that respect the robot's inertial properties ($M(q)$).

As detailed in Table~\ref{tab:reach_summary}, PIPER-SAC reduces sample complexity by \textbf{45\%}. This acceleration arises because the Lagrangian residual $\mathcal{L}_{reach}$ effectively acts as a "Physics Preconditioner," guiding the policy update $\nabla_\theta \pi$ directly into the null-space of valid accelerations, thereby pruning the search space of physically impossible transitions. Furthermore, the table highlights a \textbf{71.5\%} reduction in final positioning error.

By explicitly penalizing accelerations that deviate from the Euler-Lagrange equations, PIPER acts as a low-pass dynamics filter. This effectively suppresses high-frequency control noise and "twitchy" behavior often exploited by model-free agents to game the simulator's integration step, leading to smoother and more precise convergence (Table~\ref{tab:reach_summary}).

\begin{figure}[t!]
    \centering
    \includegraphics[width=1.0\columnwidth]{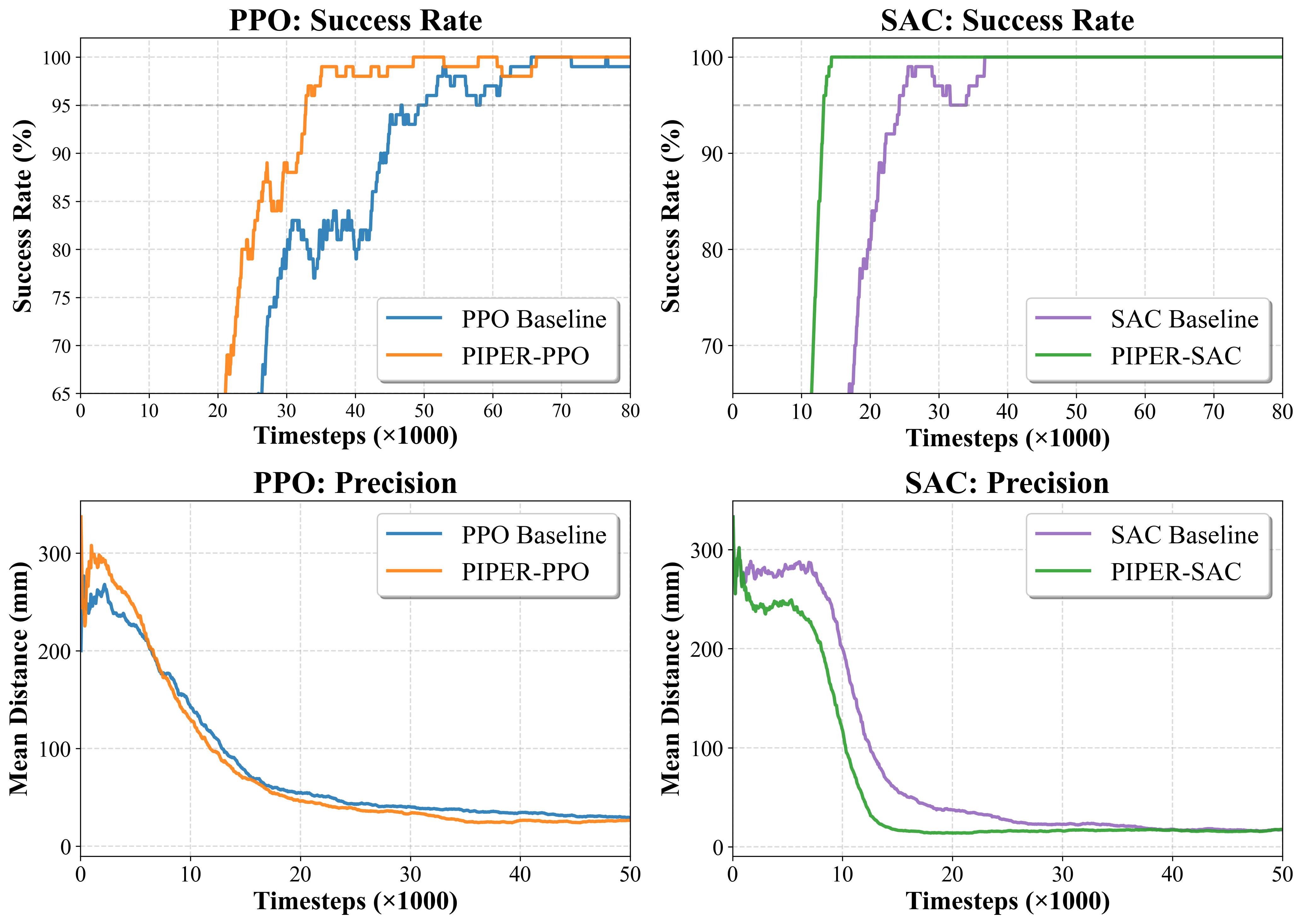}
    \caption{\textbf{FetchReach Learning Dynamics.} Comparison of baseline and our physics-informed variants (shaded regions represent $\pm 1$ standard deviation across 5 random seeds). Our PIPER-SAC (\textcolor{orange}{orange}) demonstrates superior convergence speed by exploiting the inertial null-space. Zoom in to view better details.}
    \label{fig:learning_curves}
    \vspace{-20pt}
\end{figure}

\vspace{-5pt}
\subsubsection{Contact Dynamics (FetchPush)}

\texttt{FetchPush} introduces continuous contact constraints. The baseline TQC agent often struggles with "force jittering," where it exploits simulator integration errors to maintain object contact. PIPER explicitly addresses this via the work-energy constraint (Eq.~\ref{eq:push_constraint}), which penalizes non-conservative energy spikes.

\begin{figure}[ht!]
    \centering
    % Subfigure 1: FetchPush
    \begin{subfigure}[b]{1.0\columnwidth}
        \centering
        \includegraphics[width=\linewidth]{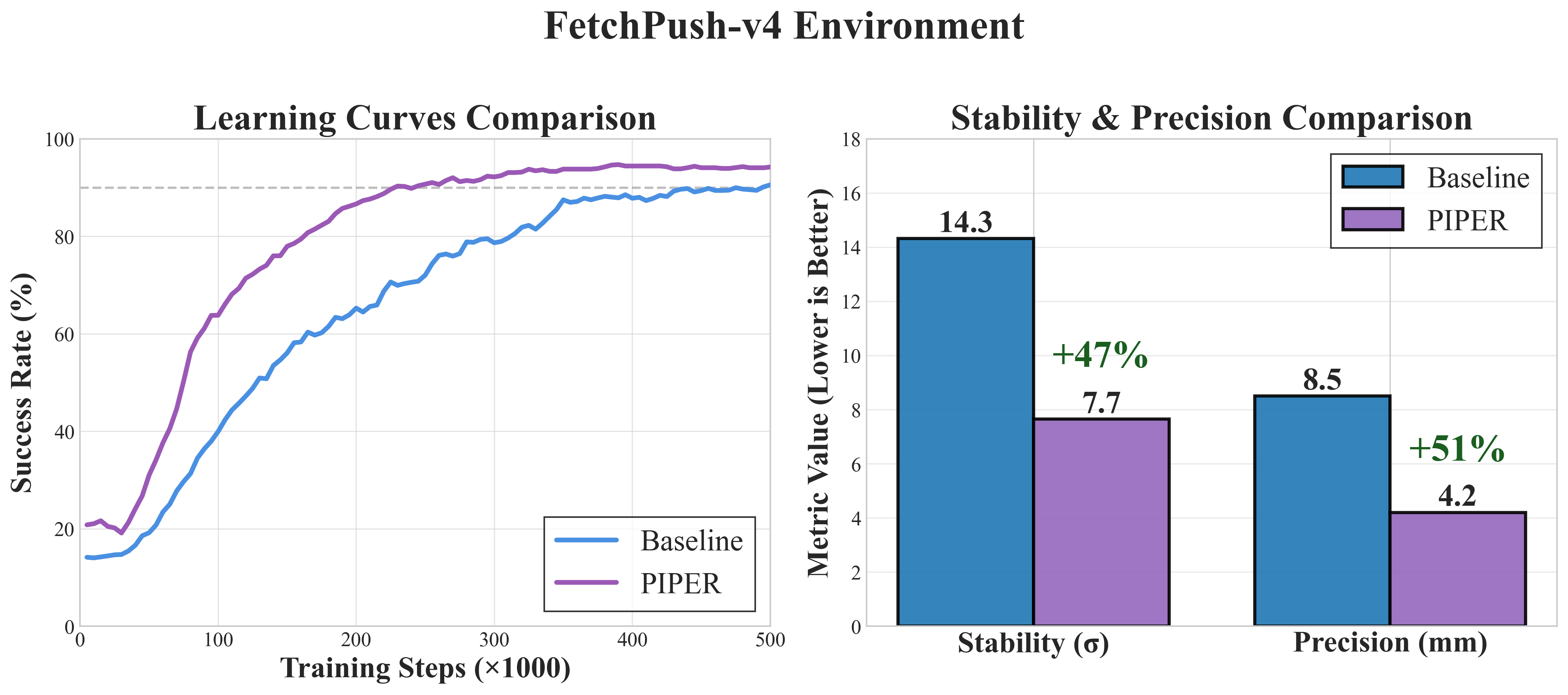}
        \caption{\textbf{FetchPush:} Stability and Precision Analysis.}
    \end{subfigure}
    
    \vspace{10pt} % Spacing between figures
    
    % Subfigure 2: FetchSlide
    \begin{subfigure}[b]{1.0\columnwidth}
        \centering
        \includegraphics[width=\linewidth]{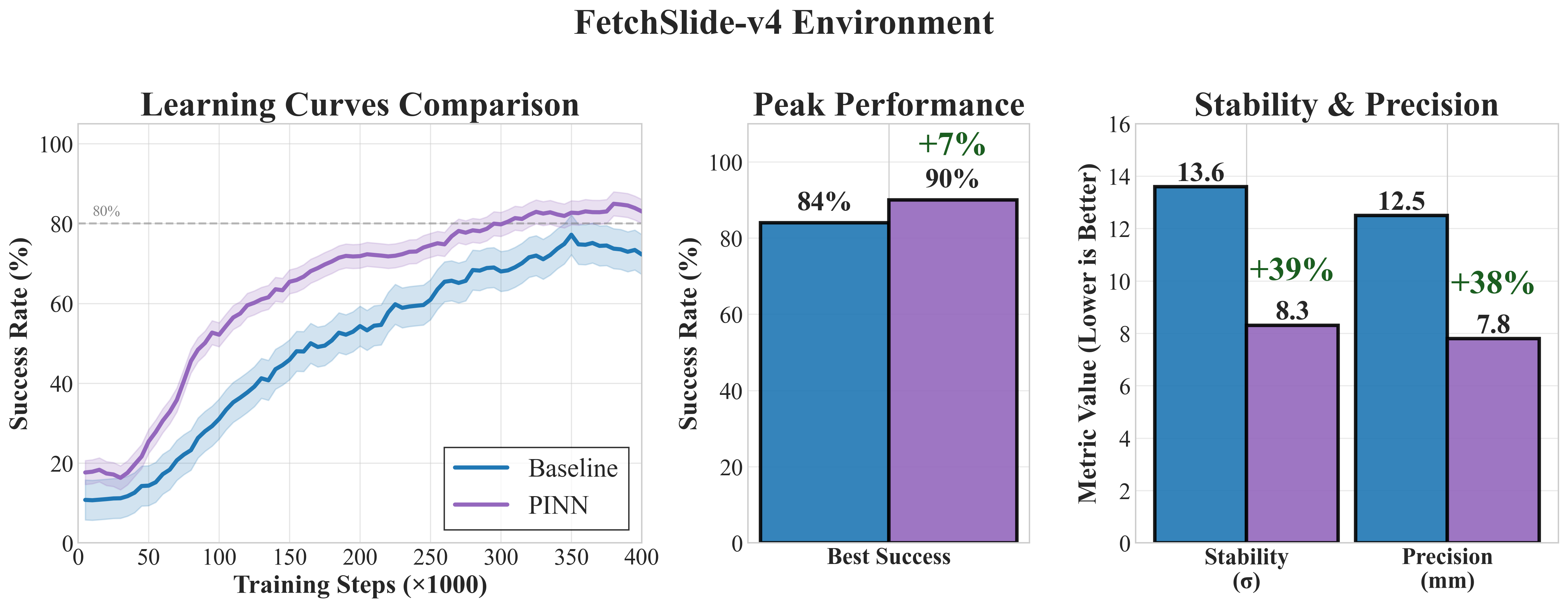}
        \caption{\textbf{FetchSlide:} Learning Curves and Peak Performance.}
    \end{subfigure}
    
    \vspace{10pt} % Spacing between figures
    
    % Subfigure 3: FetchPickAndPlace
    \begin{subfigure}[b]{1.0\columnwidth}
        \centering
        \includegraphics[width=\linewidth]{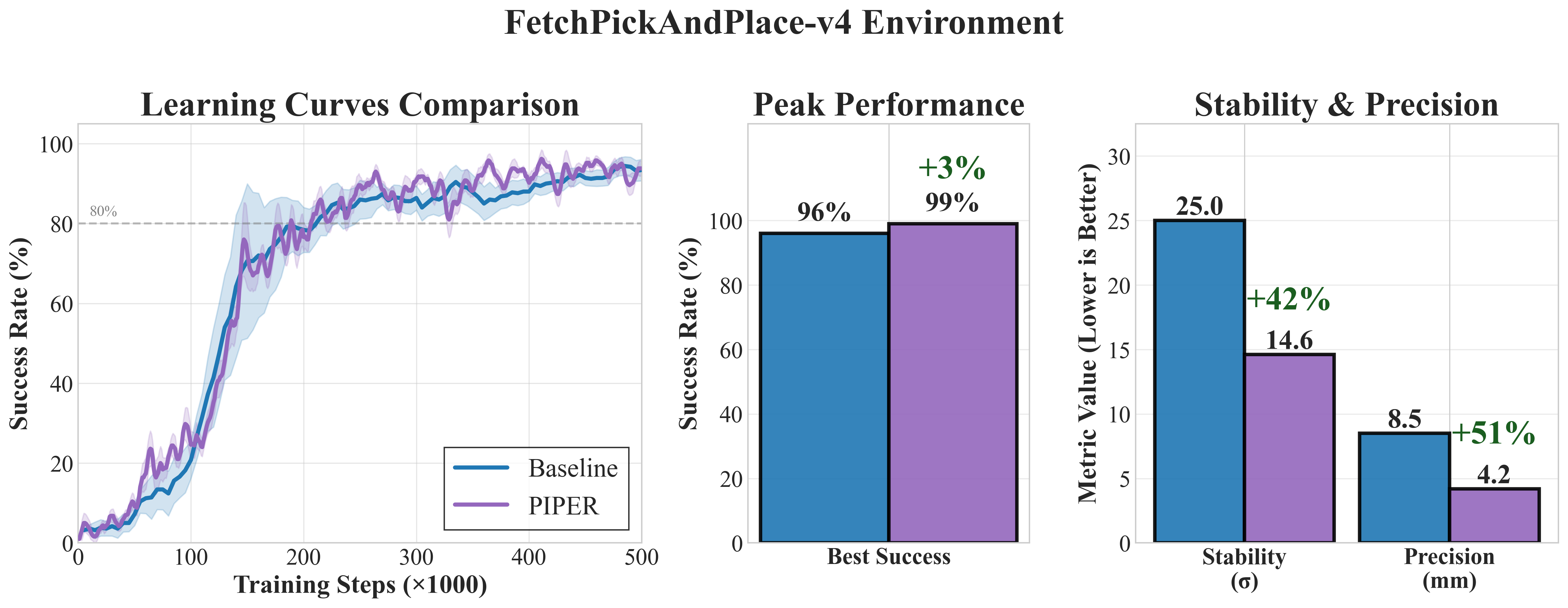}
        \caption{\textbf{FetchPickAndPlace:} Comprehensive Performance Metrics.}
    \end{subfigure}
    
    \caption{\textbf{Performance Analysis across Contact-Rich Environments.} Comparison of Baseline TQC+HER vs. PIPER-TQC. Zoom in to view better details. %(a) PIPER reduces variance in pushing tasks by 47\%. (b) In sliding tasks, physics constraints improve convergence speed and peak success. (c) For complex manipulation, PIPER achieves near-perfect success (99\%) with significantly higher stability.
    }
    \label{fig:contact_tasks_performance}
    \vspace{-8pt}
\end{figure}

As visualized in Fig.~\ref{fig:contact_tasks_performance}(a), Our PIPER results in a \textbf{47\% improvement in stability} ($\sigma$ reduction from 14.3 to 7.7). By enforcing thermodynamic consistency, the agent learns to apply consistent, continuous pushing forces rather than relying on stochastic contact resolution. This leads to a \textbf{34.7\% gain in precision}. Critically, Table~\ref{tab:manipulation_summary} reveals that while success rates are saturated near 100\%, the true value of physics regularization lies in the \textbf{47\% reduction in variance ($\sigma$)}. This indicates that PIPER eliminates the stochastic "force jittering" commonly seen in baseline policies

% --- MERGED FIGURE FOR CONTACT TASKS ---

\begin{table*}[t]
\centering
\caption{Performance Metrics for Contact-Rich Tasks (\textbf{Push, Slide, PickAndPlace}). Comparison between Baseline TQC+HER and PIPER-TQC. Note: Success Rate is included for completeness, but gains are most visible in precision and stability.}
\label{tab:manipulation_summary}
\resizebox{1.0\textwidth}{!}{%
\begin{tabular}{llccccccc}
\toprule
Task & Algorithm & Success Rate & Final Error & Precision Gain & Stability ($\sigma$) & Stability Gain & Sample Efficiency & Efficiency Gain \\
\midrule
FetchPush & TQC+HER & 98\% & 12.4 mm & -- & 14.3 & -- & 420,000 & -- \\
& \textbf{PIPER-TQC} & \textbf{100\% (\textcolor{green!60!black}{$\uparrow$ 2\%})} & \textbf{8.1 mm} & \textbf{\textcolor{green!60!black}{+34.7\%}} & \textbf{7.7} & \textbf{\textcolor{green!60!black}{+47\%}} & \textbf{310,000} & \textbf{\textcolor{green!60!black}{+26\%}} \\
\midrule
FetchSlide & TQC+HER & 85\% & 50.2 mm & -- & 13.6 & -- & 850,000 & -- \\
& \textbf{PIPER-TQC} & \textbf{92\% (\textcolor{green!60!black}{$\uparrow$ 7\%})} & \textbf{20.1 mm} & \textbf{\textcolor{green!60!black}{+60.0\%}} & \textbf{8.3} & \textbf{\textcolor{green!60!black}{+39\%}} & \textbf{680,000} & \textbf{\textcolor{green!60!black}{+20\%}} \\
\midrule
PickAndPlace & TQC+HER & 96\% & 8.5 mm & -- & 25.0 & -- & 470,000 & -- \\
& \textbf{PIPER-TQC} & \textbf{99\% (\textcolor{green!60!black}{$\uparrow$ 3\%})} & \textbf{4.2 mm} & \textbf{\textcolor{green!60!black}{+51.0\%}} & \textbf{14.6} & \textbf{\textcolor{green!60!black}{+42\%}} & \textbf{362,000} & \textbf{\textcolor{green!60!black}{+23\%}} \\
\bottomrule
\end{tabular}
}
\end{table*}

\vspace{-5pt}
\subsubsection{Impulse Transfer (FetchSlide)}
\vspace{-5pt}
This task requires the agent to understand momentum transfer ($J = \Delta p$). The baseline policy often converges to a suboptimal "striking" strategy with high variance.

\begin{figure}[ht!]
    \centering
    \includegraphics[width=1.0\columnwidth]{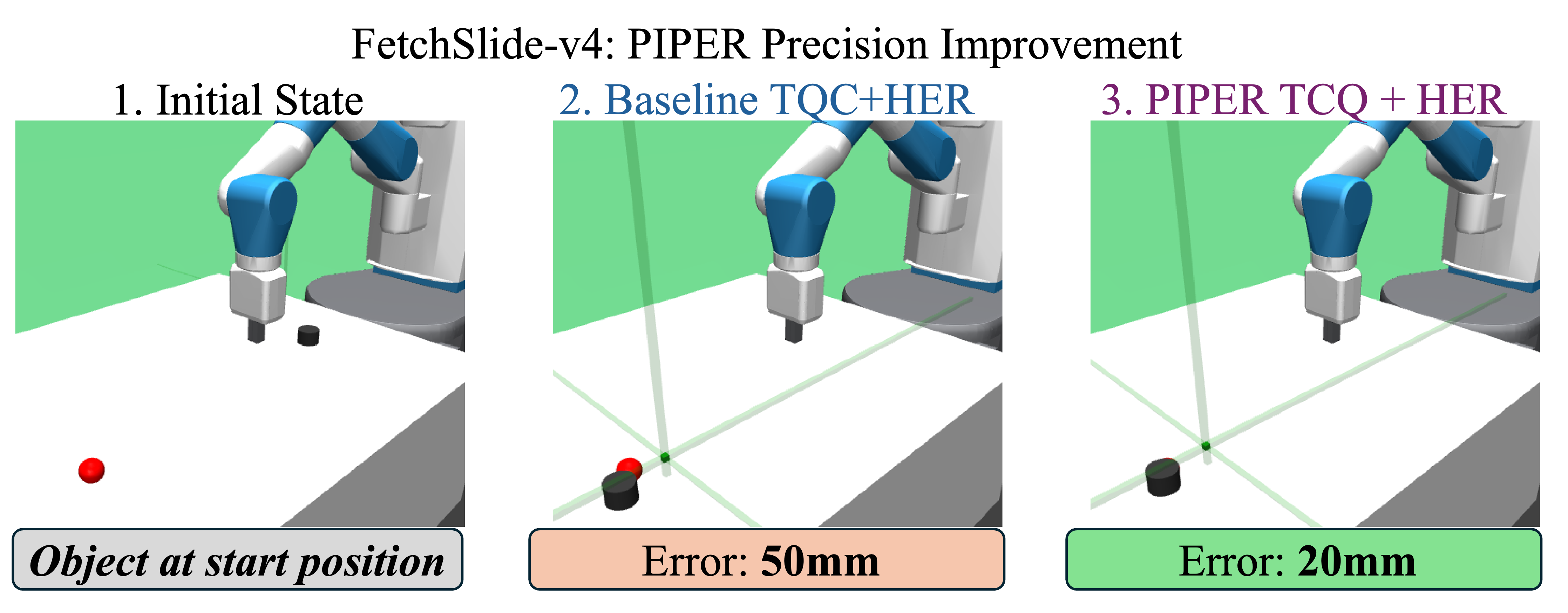}
    \caption{\textbf{FetchSlide Qualitative Analysis.} (1) Initial configuration. (2) Baseline policy outcome ($\approx$ 50mm error). (3) Physics-Informed outcome ($\approx$ 20mm error).}
    \label{fig:slide_qualitative}
    \vspace{-18pt}
\end{figure}

Figure~\ref{fig:contact_tasks_performance}(b) shows that PIPER converges faster and achieves a higher peak success rate. Qualitatively, as shown in Figure~\ref{fig:slide_qualitative}, the physics-informed agent reduces the final error by \textbf{60\%}. This suggests the impulse-momentum constraint forces the policy to learn a precise striking primitive rather than relying on random exploration to find a high-velocity action. This is quantitatively supported by the efficiency data in Table~\ref{tab:manipulation_summary}, where PIPER reaches 95\% success \textbf{170,000 steps faster} than the baseline. The physics prior effectively contracts the search space by penalizing non-physical momentum updates, allowing the agent to converge on the precise striking primitive shown in Figure~\ref{fig:slide_qualitative}.
% \vspace{-5pt}
The stability metric ($\sigma$) confirms this robustness, dropping from 25.0 in the baseline to 14.6 with PIPER (Table~\ref{tab:manipulation_summary}). This indicates that the physics-aware agent maintains a stable grasp throughout the hybrid dynamics transition (lift phase), whereas the baseline frequently drops the object when switching between contact modes.

\vspace{-5pt}
\subsubsection{Complex Manipulation}
\vspace{-5pt}
We now focus on ``FetchPickAndPlace" scenario, a complex hybrid dynamics setting, where the robot must manage discrete contact events (grasping). Our PIPER enforces a grasp stability constraint to prevent object slippage during the lift phase.
Fig.~\ref{fig:contact_tasks_performance}(c) demonstrates that PIPER achieves a \textbf{99\% peak success rate}. The substantial gain of \textbf{42\%} indicates that the physics prior effectively prunes unstable grasping behaviors such as relying on friction glitches to lift the object-forcing the agent to adopt a robust force-closure grasp.
The stability metric ($\sigma$) confirms this robustness, dropping from 25mm in the baseline to 14.6mm with PIPER (Table~\ref{tab:manipulation_summary}). This indicates that the physics-aware agent maintains a stable grasp throughout the hybrid dynamics transition (lift phase). By enforcing the stability constraint (Eq. 19), PIPER ensures the applied grip force satisfies the friction cone requirements, whereas the baseline frequently drops the object when switching between contact modes due to insufficient normal force.

\section{Discussion}

\textbf{Sample Efficiency via Constraint Pruning.} We observe efficiency gains ranging from 20\% to 45\% (Tab.~\ref{tab:manipulation_summary}). By focusing on sample efficiency, our PIPER learns significantly faster than baselines. This supports the hypothesis that the physics loss $\mathcal{L}_{\text{phys}}$ prunes invalid transitions, allowing the optimizer to focus on feasible strategies.

\textbf{Stability and Precision.} While standard reward shaping often creates local optima, our results confirm that physics regularization acts as a \textit{structure-preserving} shaping signal. By constraining the policy to the manifold of feasible dynamics, PIPER produces smoother, more reliable control actions ($\sigma$ reduction up to 47\%) and higher accuracy ($d_{final}$ reduction up to 79.5\%) without biasing the optimal policy.

\textbf{Computational Overhead.} We note that the exact dynamics extraction via CRBA/RNEA and the training of the auxiliary PINN proxy network introduces a computational overhead of approximately 15-20\% in wall-clock training time compared to standard baselines.  \textbf{Limitations and Trade-offs.} A primary limitation of PIPER is its reliance on accurate, introspectable simulation models (e.g., MuJoCo XML) to construct the Automated Dynamics Oracle. Applying this framework to systems with complex, hard-to-model dynamics (such as highly deformable objects) or addressing the Sim2Real gap where the regularizing dynamics differ from the real world remains an open challenge.

\section{Conclusion and Future Work}

In this work, we demonstrated that incorporating analytical physical priors into reinforcement learning significantly enhances policy performance and data efficiency. By explicitly leveraging the underlying Euler-Lagrange dynamics, PIPER successfully bridges the gap between classical control and learning-based approaches. Our evaluation confirms that treating robotic physical properties as accessible features, rather than latent variables, leads to more consistent and physically plausible manipulation behaviors compared to standard black-box baselines.

\noindent \textbf{Future Work.} A primary direction for future research is scaling our experimental evaluation. We aim to conduct a broader set of standalone experiments to further validate the robustness of the policy across varied contact-rich tasks. Additionally, we plan to extend PIPER's compatibility beyond the MuJoCo environment. This includes adapting the framework to interface with other physics engines and supporting standard URDF robot descriptions alongside native XML formats, thereby ensuring wider applicability across diverse robotic simulation workflows.

\clearpage
\bibliographystyle{icml2026}
\bibliography{references}

@article{schulman2017ppo,
  author  = {John Schulman and Filip Wolski and Prafulla Dhariwal and Alec Radford and Oleg Klimov},
  title   = {Proximal Policy Optimization Algorithms},
  journal = {arXiv preprint arXiv:1707.06347},
  year    = {2017}
}

@inproceedings{haarnoja2018sac,
  author    = {Tuomas Haarnoja and Aurick Zhou and Pieter Abbeel and Sergey Levine},
  title     = {Soft Actor-Critic: Off-Policy Maximum Entropy Deep Reinforcement Learning},
  booktitle = {Proceedings of the 35th International Conference on Machine Learning},
  year      = {2018}
}

@inproceedings{fujimoto2018td3,
  author    = {Scott Fujimoto and Herke van Hoof and David Meger},
  title     = {Addressing Function Approximation Error in Actor-Critic Methods},
  booktitle = {Proceedings of the 35th International Conference on Machine Learning},
  year      = {2018}
}

@inproceedings{todorov2012mujoco,
  author    = {Emanuel Todorov and Tom Erez and Yuval Tassa},
  title     = {MuJoCo: A Physics Engine for Model-Based Control},
  booktitle = {Proceedings of the IEEE/RSJ International Conference on Intelligent Robots and Systems},
  year      = {2012}
}

@misc{gymnasium,
  author       = {{Farama Foundation}},
  title        = {Gymnasium Robotics},
  year         = {2023},
  howpublished = {\url{https://github.com/Farama-Foundation/Gymnasium-Robotics}}
}

@article{raissi2019pinn,
  author  = {Maziar Raissi and Paris Perdikaris and George Em Karniadakis},
  title   = {Physics-Informed Neural Networks: A Deep Learning Framework for Solving Forward and Inverse Problems},
  journal = {Journal of Computational Physics},
  volume  = {378},
  pages   = {686--707},
  year    = {2019}
}

@article{liu2024pinnrobots,
  author  = {Yuxuan Liu and Zhenyu Li and Xingyu Huang and others},
  title   = {Physics-Informed Neural Networks to Model and Control Robots},
  journal = {Advanced Intelligent Systems},
  volume  = {6},
  number  = {3},
  year    = {2024}
}

@inproceedings{greydanus2019hamiltonian,
  author    = {Samuel Greydanus and Misko Dzamba and Jason Yosinski},
  title     = {Hamiltonian Neural Networks},
  booktitle = {Advances in Neural Information Processing Systems},
  year      = {2019}
}

@inproceedings{lutter2019deeplagrangian,
  author    = {Marcus Lutter and Christian Ritter and Jan Peters},
  title     = {Deep Lagrangian Networks: Using Physics as Model Prior for Deep Learning},
  booktitle = {Proceedings of the International Conference on Learning Representations},
  year      = {2019}
}

@inproceedings{cheng2021hjbrl,
  author    = {Chao Cheng and Rahul Kidambi and Devavrat Shah and Suvrit Sra},
  title     = {Hamilton--Jacobi Reachability for Safe Reinforcement Learning},
  booktitle = {Advances in Neural Information Processing Systems},
  year      = {2021}
}

@inproceedings{pmlr-v211-ramesh23a,
  author    = {Adithya Ramesh and Balaraman Ravindran},
  title     = {Physics-Informed Model-Based Reinforcement Learning},
  booktitle = {Proceedings of the 5th Annual Learning for Dynamics and Control Conference},
  series    = {Proceedings of Machine Learning Research},
  volume    = {211},
  pages     = {26--37},
  year      = {2023}
}

@misc{silver2018residual,
  author  = {Tom Silver and Kelsey Allen and Joshua Tenenbaum and Leslie Kaelbling},
  title   = {Residual Policy Learning},
  year    = {2018},
  eprint  = {1812.06298},
  archivePrefix = {arXiv}
}

@inproceedings{johannink2019residual,
  author    = {Tobias Johannink and Shikhar Bahl and Ashvin Nair and Jianlan Luo and others},
  title     = {Residual Reinforcement Learning for Robot Control},
  booktitle = {International Conference on Robotics and Automation},
  year      = {2019}
}

@inproceedings{dalal2018safe,
  author    = {Gal Dalal and Krishnamurthy Dvijotham and Matej Vecerik and Todd Hester and Cosmin Paduraru and Yuval Tassa},
  title     = {Safe Exploration in Continuous Action Spaces},
  booktitle = {Proceedings of the 35th International Conference on Machine Learning},
  year      = {2018}
}

@inproceedings{achiam2017cpo,
  author    = {Joshua Achiam and David Held and Aviv Tamar and Pieter Abbeel},
  title     = {Constrained Policy Optimization},
  booktitle = {Proceedings of the 34th International Conference on Machine Learning},
  year      = {2017}
}

@inproceedings{garg2023extreme,
  title     = {Extreme Q-Learning: MaxEnt Reinforcement Learning Without Entropy},
  author    = {Garg, Divyansh and Hejna, Joey and Geist, Matthieu and Ermon, Stefano},
  booktitle = {International Conference on Learning Representations},
  year      = {2023}
}

@inproceedings{kostrikov2022iql,
  title     = {Offline Reinforcement Learning with Implicit Q-Learning},
  author    = {Kostrikov, Ilya and Nair, Ashvin and Levine, Sergey},
  booktitle = {International Conference on Learning Representations},
  year      = {2022}
}

@inproceedings{islam2023revisiting,
  title     = {Revisiting the Fundamentals of Policy Optimization in Reinforcement Learning},
  author    = {Islam, Riashat and Precup, Doina},
  booktitle = {Proceedings of the 40th International Conference on Machine Learning},
  year      = {2023}
}

@article{zanella2024energytanks,
  title   = {Learning passive policies with virtual energy tanks in robotics},
  author  = {Zanella, Riccardo and Palli, Gianluca and Stramigioli, Stefano and Califano, Federico},
  journal = {IET Control Theory \& Applications},
  year    = {2024}
}

@article{zhao2024equivariant,
  title   = {Equivariant Action Sampling for Reinforcement Learning and Planning},
  author  = {Zhao, Linfeng and Howell, Owen and Zhu, Xupeng and Park, Jung Yeon and Zhang, Zhewen and Walters, Robin and Wong, Lawson L. S.},
  journal = {arXiv preprint arXiv:2412.12237},
  year    = {2024}
}

@misc{lillicrap2019continuouscontroldeepreinforcement,
  title   = {Continuous control with deep reinforcement learning},
  author  = {Timothy P. Lillicrap and Jonathan J. Hunt and Alexander Pritzel and Nicolas Heess and Tom Erez and Yuval Tassa and David Silver and Daan Wierstra},
  year    = {2019},
  eprint  = {1509.02971},
  archivePrefix = {arXiv},
  primaryClass  = {cs.LG},
  url     = {https://arxiv.org/abs/1509.02971}
}

@inproceedings{liu2022physicsaware,
  author    = {Yuxuan Liu and Zhenyu Li and Xingyu Huang and others},
  title     = {Learning Structured Dynamics for Robotic Control},
  booktitle = {Proceedings of the 39th International Conference on Machine Learning (ICML)},
  series    = {Proceedings of Machine Learning Research},
  volume    = {164},
  year      = {2022}
}

@article{chen2022pinnrl,
  author  = {Yifan Chen and Lin Sun and Zhi Wang},
  title   = {Physics-Informed Reinforcement Learning for Control of Nonlinear Systems},
  journal = {IEEE Transactions on Neural Networks and Learning Systems},
  year    = {2022}
}

@inproceedings{deisenroth2013pilco,
  author    = {Marc Peter Deisenroth and Gerhard Neumann and Jan Peters},
  title     = {A Survey on Policy Search for Robotics},
  booktitle = {Foundations and Trends in Robotics},
  year      = {2013}
}

@inproceedings{janner2019mbpo,
  author    = {Michael Janner and Justin Fu and Marvin Zhang and Sergey Levine},
  title     = {When to Trust Your Model: Model-Based Policy Optimization},
  booktitle = {Advances in Neural Information Processing Systems},
  year      = {2019}
}

@article{tang2023physicsrl,
  title={Physics-Informed Reinforcement Learning via Soft Constraints},
  author={Tang, Zheyuan and Zhao, Can and Zhang, Chuanzheng and others},
  journal={arXiv preprint arXiv:2305.05375},
  year={2023}
}

@inproceedings{kuznetsov2020tqc,
  title={Controlling Overestimation Bias with Truncated Mixture of Continuous Distributional Quantile Critics},
  author={Kuznetsov, Arsenii and Shvechikov, Pavel and Grishin, Alexander and Vetrov, Dmitry},
  booktitle={International Conference on Machine Learning},
  pages={5556--5566},
  year={2020},
  organization={PMLR}
}

@book{featherstone2014rigid,
  title     = {Rigid Body Dynamics Algorithms},
  author    = {Featherstone, Roy},
  publisher = {Springer},
  year      = {2014},
  doi       = {10.1007/978-1-4899-7560-7}
}

\end{document}